\pdfoutput=1

\documentclass[11pt]{article}

\usepackage[]{acl}

\usepackage{times}
\usepackage{latexsym}
\usepackage{bm}
\usepackage{hyperref}
\usepackage{color,colortbl}
\usepackage{stfloats}
\usepackage{enumitem}
\usepackage{booktabs}
\usepackage{nccmath}
\usepackage{multirow}
\usepackage{subfigure}
\usepackage{latexsym}
\usepackage{todonotes}
\usepackage{makecell}
\usepackage{dashrule}
\usepackage{amssymb}
\usepackage{bbding}

\usepackage{graphicx}

\definecolor{LightGray}{gray}{0.9}
\usepackage{soul}

\usepackage[T1]{fontenc}

\usepackage[utf8]{inputenc}

\usepackage{microtype}

\title{
Multi-Figurative Language Generation}

\author{Huiyuan Lai \and Malvina Nissim \\
    Center for Language and Cognition (CLCG) \\
    University of Groningen / The Netherlands\\
  \texttt{\{h.lai, m.nissim\}@rug.nl} \\
}

\begin{document}
\maketitle
\begin{abstract}
Figurative language generation is the task of reformulating a given text in the desired figure of speech while still being faithful to the original context. 
We take the first step towards multi-figurative language modelling by providing a benchmark for the automatic generation of five common figurative forms in English. We train \textbf{mFLAG} employing a scheme for multi-figurative language pre-training on top of BART, and a mechanism for injecting the target figurative information into the encoder; this enables the generation of text with the target figurative form from another figurative form without parallel figurative-figurative sentence pairs. Our approach outperforms all strong baselines. We also offer some qualitative analysis and reflections on the relationship between the different figures of speech.
\end{abstract}

\section{Introduction}
Figurative language is commonly used in speaking and writing to accomplish a constellation of communicative goals~\citep{richard-etal-1994}. Figures of speech, such as metaphors, or idiomatic expressions, can make an expression stand out by making it more interesting and captivating, and can evoke stronger 
emotions than more factual, literal phrases thereby making the text more engaging. 

Automatic figurative language generation has received growing attention with the progress of neural networks, especially the emergence of large pre-trained models~\citep{raffel-etal-2020, lewis-etal-2020-bart}. We see there are two core values for this task: (i) computational approaches can be employed to provide a better understanding of linguistic phenomena and more specifically in this case different figures of speech; (ii) we can explore how much models can handle creativity and devise ways to employ them in the support of creative writing, so as to yield more varied and human-like generated text, 
including in the context of machine translation~\citep{Ana-antonio-2022}.

There are many related tasks that have been proposed and studied by NLP researchers, including the generation of hyperbole~\citep{tian-etal-2021-hypogen-hyperbole, zhang2021mover}, idiom~\citep{zhou2021solving}, sarcasm~\citep{zhu2019neural, chakrabarty-etal-2020-r}, metaphor~\citep{Abe-etal-2006-acm, stowe-etal-2021-metaphor}, and simile~\citep{chakrabarty-etal-2020-generating, zhang-etal-2021-writing}. Table~\ref{tab:examples} shows examples of figurative language generation from literal texts.

Previous works focus on modelling single figurative forms, generally rewriting a literal sentence into one with a specific figure of speech. 
This results in having to train separate models, one for each figure of speech, and in not exploiting knowledge transfer across figurative forms. However, since different figures of speech can share some features related to non-literality, and a  text may also contain and combine multiple figures of speech at the same time, it is possible that substantial knowledge gains can be transferred from one figure to another. Moreover, the generation between different figures of speech (e.g.\ generating an idiomatic text from the hyperbolic one) is under-explored. 

\begin{table}[t]
\setlength{\tabcolsep}{4pt}
\resizebox{\linewidth}{!}{%
\centering
\footnotesize
\begin{tabular}{ll}
\toprule[1pt]
 \textbf{Forms} & \textbf{Sentences}\\
 \hline
 Literal & Old Mr. Smith has been teaching here for a very long time.\\
 Hyperbole     & Old Mr. Smith has been teaching here since the Stone Age.\\
 \hline
 Literal & My niece will babysit for you for a little bit of money.\\
 Idiom         & My niece will babysit for you for pin money.\\
 \hline
 Literal & I hate it when they run the same commercial twice in a row.
\\
 Sarcasm         & I love when they run the same commercial twice in a row.\\
 \hline
 Literal & He remembers a road of my broken works.\\
 Metaphor      & He made a road of my broken works.\\
 \hline
 Literal & You can publish the whole thing old.\\
 Simile        & You can publish the whole thing like a diary.\\
\bottomrule[1pt]
\end{tabular}}
\caption{\label{tab:examples}
Examples of figurative language generation from literal texts.
}
\end{table}

In this work we suggest to model multiple figures of speech jointly, with the ultimate goal of having a single model that can handle the generation of multiple figurative forms from both literal and figurative inputs.

Intuitively, multi-task learning~\citep{collobert-etal-2008-multi} and the usage of a domain label~\citep{kobus-etal-2017-domain} could be a good method for multi-figurative language modelling, adding a special token to the beginning of the sentence to guide text generation. Such a method requires parallel data (i.e.\ aligned texts with the same context but different figures of speech) for training; this is usually unavailable, especially between different figures of speech, and costly to produce.

We rely on existing parallel data between literal sentences and single figures of speech and propose mFLAG (\textbf{M}ulti-\textbf{F}igurative \textbf{La}nguage \textbf{G}eneration), 
an approach which is applicable to the generation between different forms, both literal and figurative. In a nutshell, mFLAG is trained in two stages, in both of which we also exploit the contribution of generic paraphrase data: (i) a specifically designed pre-training for multi-figurative language, where a special label is added at the beginning of each sentence to indicate its figure of speech; (ii) a supervised training where the parallel literal-figurative sentence pairs for all figurative languages are combined to achieve multi-figurative language generation. For (ii), we introduce an innovative mechanism that allows the form labels to leak their own figurative information into the input embedding, thus guiding the encoder to represent the source sentence. This mechanism makes it possible to generate between different figures of speech without parallel figurative-figurative data. For comparison, and to allow for wider flexibility in generation choices as well as linguistic analysis, we also use the literal form corresponding to each figure of speech, which is available through the separate parallel datasets, as pivot to run figurative-to-figurative transformation. We expect that with the direct figurative-figurative transformation the source figurative form might still be maintained in the generated sentence, with the addition of the target figurative form, while this should not be the case when using the literal form as pivot.

\paragraph{Contributions} Considering five common figures of speech in English, (i) we propose a novel task of multi-figurative language generation, and explore the potential of its computational modelling; (ii) we introduce a pre-training scheme for multi-figurative language modelling, which boosts performance substantially by leveraging paraphrase data and cross-figurative language knowledge transfer;
(iii) we design a mechanism for injecting the desired figurative information into the encoder to achieve the generation between different figures of speech without parallel figurative-figurative sentence pairs; this mechanisms could be applied to other tasks, too; (iv) we compare  figurative-figurative and figurative-literal-figurative generation, thereby assessing the feasibility, the limits, and the characteristics of direct multi-figurative language generation; and (v) we provide a benchmark for multi-figurative language generation, which can hopefully foster the progress of figurative language processing. \footnote{Data, code, and model are available at \url{https://github.com/laihuiyuan/mflag}.}

\section{Background}
Transforming text involving a figure of speech, either in source or in target or both, is closely related to three other NLP tasks, namely paraphrasing, text style transfer, and figurative language detection. We discuss relevant background on such tasks, and why and how they play a role in our work. 

\paragraph{Paraphrasing}
Paraphrasing is the task of generating a text semantically (almost) identical to a given input, but with variations in wording or syntax~\citep{prakash-etal-2016-neural, cao-etal-2017-joint}. The large amount of parallel paraphrase data available can be used to teach models a general rewriting task in the context of various downstream NLP tasks, such as semantic parsing~\citep{berant-liang-2014-semantic}, machine translation~\citep{callison-burch-etal-2006-improved}, question answering~\citep{dong-etal-2017-learning}, and text style transfer~\citep{lai-etal-2021-generic}. As figurative generation can be viewed as a special paraphrasing task, where texts are expected to include specific figurative forms, we also leverage paraphrase data for figurative generation modelling.

\paragraph{Text Style Transfer}
The goal of text style transfer is to transform a given text of one style into another while preserving the style-independent content. A common task, for example, is formality transfer, where an informal sentence is turned into formal, or viceversa~\citep{rao-tetreault-2018}.
Generally speaking, both text style transfer and figurative language generation aim to achieve the generation of text with specific attributes. 
Regarding sentence changes, for text style transfer, often multiple parts of the sentence might be modified at the same time, such as capitalization at the beginning of the sentence, punctuation at the end, and some phrasing in the middle. 
Figurative language generation, instead,  often concerns the rewriting of some specific expressions, while other (possibly large) portions of the input sentence could be retained~\citep{zhou2021solving}.
Also, in figurative language generation, the original figurative form could be still present in the transformed sentence, while text style transfer aims to alter the original style fully.

It should also be pointed out that addressing multi-figurative language generation is particularly challenging since not all figures of speech considered require the same kinds of alterations in text.

\paragraph{Figurative Language Detection}
Most past work on figurative language processing focuses on detection rather than generation.
The detection of figurative language generally involves two levels: sentence-level and word-level. At sentence-level, the task is usually formulated as a binary classification problem, namely automatically detecting whether a given sentence is literal or non-literal~\citep{troiano-etal-2018-computational}. At word-level, the task is concerned with identifying the
exact words within a sentence which trigger the figurative reading~\citep{beigman-klebanov-etal-2016-semantic, mao-etal-2018-word}. This task is a crucial component in retrieval-based approaches to figurative language generation, which usually require first the identification of triggering words in a sentence, followed then by other operations such as replacement and generation (see next paragraph.)

\paragraph{Figurative Language Generation}
Early work on figurative language generation is mainly template-based. ~\citet{Abe-etal-2006-acm} employ simple expressions ``A is like B'' for metaphor generation.~\citet{veale-2016-round} use template-like structures to generate metaphoric tweets. These methods usually lack the flexibility to cope with the variability intrinsic to (creative) natural language. In recent years, 
figurative language modelling has mostly shifted to neural-based end-to-end approaches, showing good degrees of creativity, for example in the generation of puns and metaphors~\citep{yu-etal-2018-neural, yu-wan-2019-avoid}. To provide better explainability,~\citet{zhou2021solving} propose a neural-based pipeline for idiom generation that contains three explicit steps: retrieve, extract, and generate. Most recently, and as in most NLP tasks, impressive results for figurative language generation have been achieved leveraging pre-trained models. For example, ~\citet{stowe-etal-2021-exploring} and ~\citet{chakrabarty-etal-2021-mermaid} successfully generate metaphors fine-tuning T5~\citep{raffel-etal-2020} and BART~\citep{lewis-etal-2020-bart}, respectively. Fine-tuning BART is successful for the generation of simile~\citep{chakrabarty-etal-2020-generating}, and hyperbole~\citep{zhang2021mover}, too.~\citet{stowe-etal-2021-metaphor} also propose to control the metaphor generation process by encoding conceptual mappings in the form of FrameNet frames. All these works focus on single figurative forms, modelling generation between literal and figurative. Instead, while still leveraging parallel literal-figurative data for single forms, we aim to model multiple figures of speech jointly thereby also generative between different figurative forms.

\begin{table}[t]
\setlength{\tabcolsep}{4pt}
\resizebox{\linewidth}{!}{%
\centering
\footnotesize
\begin{tabular}{llrrr}
\toprule[1pt]
 \textbf{Forms} & \textbf{Task} & \textbf{Train} & \textbf{Valid} & \textbf{Test}\\
 \hline
 Hyperbole & Literal Form$\leftrightarrow$Hyperbole & 509(+668) & 50 & 150\\
 Idiom     & Literal Form$\leftrightarrow$Idiom     & 3,784 & 876 & 876\\
 Sarcasm     & Literal Form$\leftrightarrow$Sarcasm     & 16,762 & 1,500 & 1,470\\
 Metaphor  & Literal Form$\leftrightarrow$Metaphor  & 118,807 & 6,254 & 150\\
 Simile    & Literal Form$\leftrightarrow$Simile & 82,687 & 5,145 & 150\\
\bottomrule[1pt]
\end{tabular}}
\caption{\label{tab:data-statistics}
Dataset statistics.
}
\end{table}

\section{Task and Dataset}
\label{sec:data}
We define the task of figurative language generation as the transformation of a text written in (or with) a given form (literal or figurative) to a text in (or containing) another form, while preserving the original general context. 

We use five existing datasets for the figures of speech we consider in this paper; Table~\ref{tab:data-statistics} shows sizes and splits.

\begin{itemize}[leftmargin=*]
\itemsep 0in

\item \textbf{Hyperbole}
~\citet{troiano-etal-2018-computational} introduce HYPO, a corpus of 709 hyperbolic sentences with their non-hyperbolic formulations. We boost this small dataset with some automatically obtained pairs. We fine-tune BART with HYPO, and use this model to transform into literal the hyperbolic texts contained in the non-parallel dataset HYPO-Red~\citep{tian-etal-2021-hypogen-hyperbole}. We then select literal generations with a low hyperbolic score $\sigma$ as predicted by a binary classifier based on BERT~\citep{devlin-etal-2019-bert} trained on HYPO, for an additional 668 training pairs.\footnote{Generated literal texts with $\sigma < 0.5$ are selected.}

\item \textbf{Idiom}
~\citet{zhou2021solving} use the existing MAGPIE corpus~\citep{haagsma-etal-2020-magpie} to create a parallel dataset of literal and idiomatic pairs.

\item \textbf{Sarcasm}
~\citet{peled-reichart-2017-sarcasm} release a dataset of 3,000 pairs of sarcastic tweets each augmented with five interpretations. We complement this by adding to the training set 4,762 sentence pairs from a sarcasm dataset~\citep{ghosh-etal-2020-interpreting}.

\item \textbf{Metaphor}
~\citet{stowe-etal-2021-metaphor} build a literal-metaphor dataset exploiting the Gutenberg Poetry corpus~\citep{Jacobs2018TheGE}: metaphoric verbs are identified, masked, and eventually replaced with infilling from a language model.

\item \textbf{Simile}
\citet{chakrabarty-etal-2020-generating} automatically collect a set of self-labelled similes via distant supervision, using the phrase \textit{like a}; similes are converted into their literal versions leveraging the structured common sense knowledge obtained from COMET~\citep{bosselut-etal-2019-comet}.
\end{itemize}

\paragraph{Pre-Training Data}
Given that figurative generation is a special paraphrasing task, we use the available paraphrase data from PARABANK~2~\citep{hu-etal-2019-large} for multi-figurative language modelling, but only selecting more relevant pairs for the pre-training phase. To do so, we fine-tune BERT with the above figurative data to obtain five binary classifiers (each one literal vs figurative). With them, we do figurative language detection on paraphrase data, and only retain pairs where the probability that the source and target sentences are in literal form and figurative form, respectively, is greater than a threshold $\sigma$.\footnote{More details about the pre-training data for each figure of speech are in Appendix~\ref{app:pre-training-data}.}

\begin{figure*}[t]
    \begin{minipage}[t]{0.5\linewidth}
    \subfigure[Multi-figurative language denoising pre-training and fine-tuning.]{
      \includegraphics[scale=0.52]{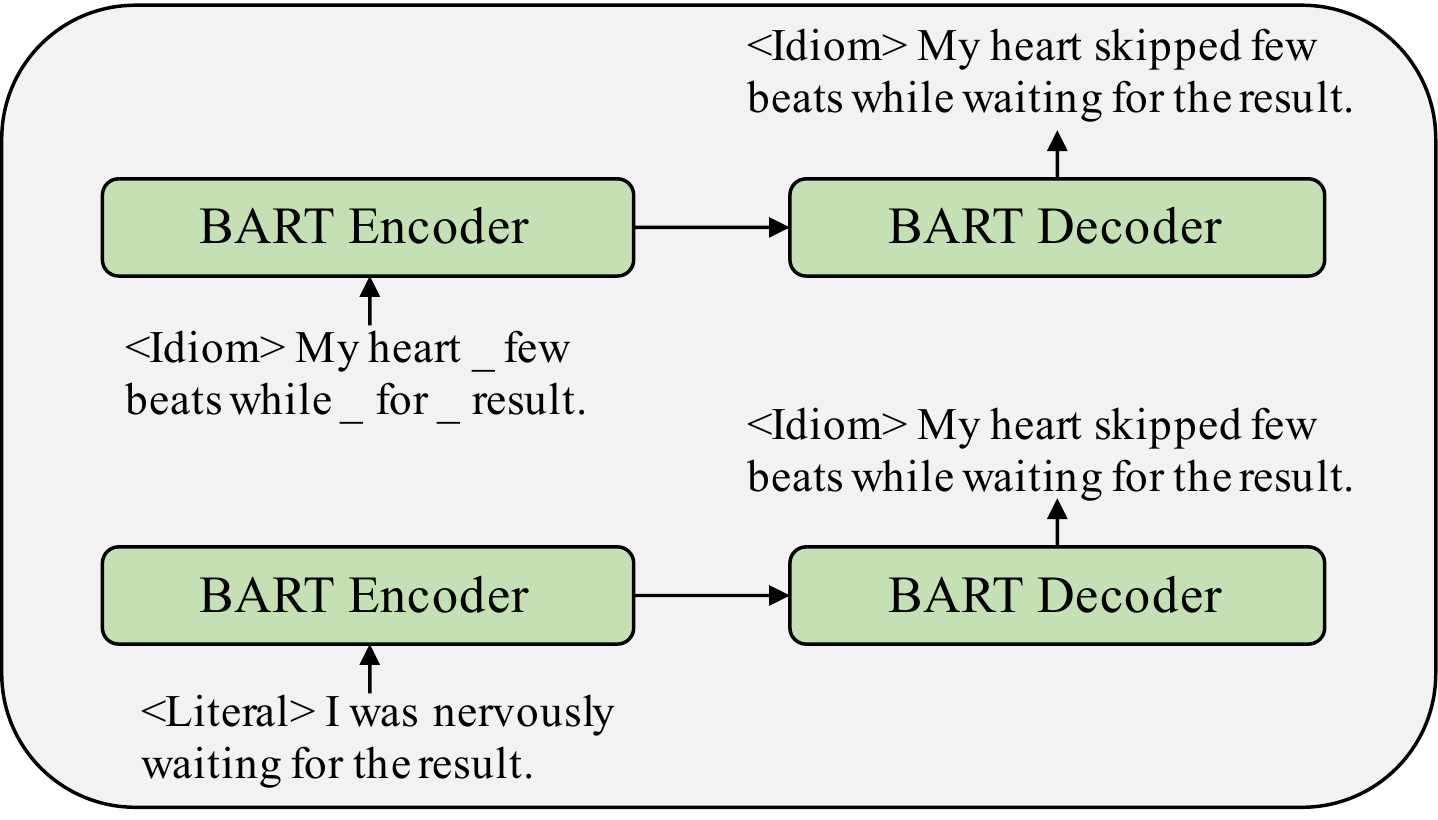}
      \label{fig:overview-bart}
    }
    \end{minipage}
    \begin{minipage}[t]{0.5\linewidth}
    \subfigure[An overview of the mechanism for injecting the figurative information into the Encoder.]{
        \includegraphics[scale=0.525]{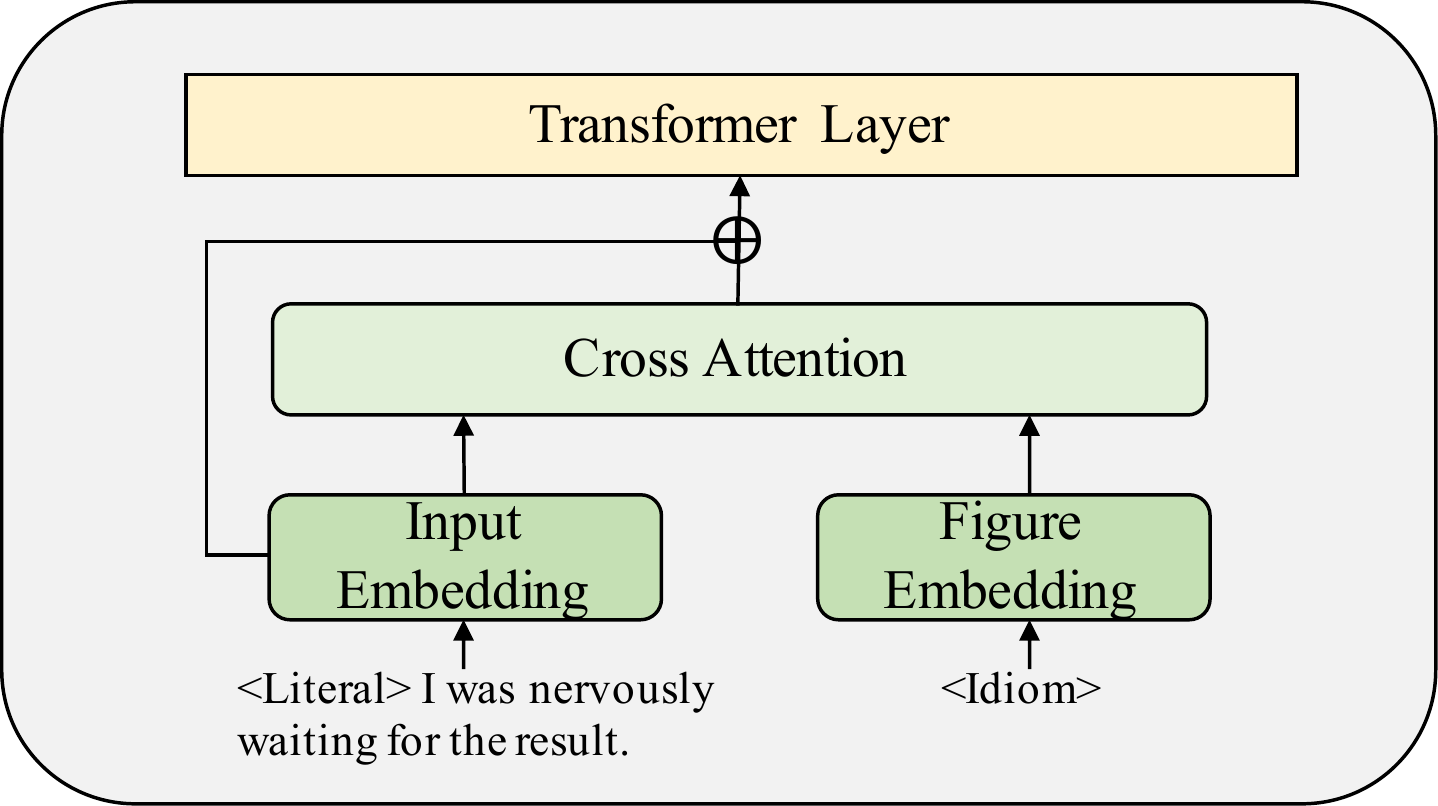} 
        \label{fig:overview-encoder}
    }
    \end{minipage}
    \caption{Overview of multi-figurative language modelling. In~\ref{fig:overview-bart}, there is the framework for our multi-figurative language denoising pre-training (top) where word masking as the injected noise, and fine-tuning on downstream task of figurative language generation (down); in~\ref{fig:overview-encoder}, the figurative information is injected into the encoder using cross-attention and residual learning. }
    \label{fig:overview}
\end{figure*}

\section{Multi-figurative Language Modelling}
We propose an approach to model multi-figurative language on top of the large pre-trained sequence-to-sequence model BART~\citep{lewis-etal-2020-bart}, by performing further, figurative language-specific pre-training, and then fine-tuning.

BART is a seq2seq model trained as a denoising autoencoder, and to reconstruct the original text $T$ given $g(T)$ where $g$ is a noising function that is used to corrupt text:
\begin{equation}
\label{eq:denoising}
    L_{\theta} = -\sum \log (T\mid g(T); \theta)
\end{equation}
with $\theta$ being the parameters of BART.

\subsection{Multi-figurative Language Pre-training}
We further pre-train BART for multi-figurative language modelling with a procedure that creates one model capable of modelling multiple figurative languages at once, so that (i) only one model needs to be maintained, and (ii) the model can benefit from cross-figurative knowledge transfer.

 Inspired by~\citet{tang2020multilingual}, we use a special token as a prefix in both the source and target text. That is, the text format is [form code] $T$ [eos] with $T$ being the text and the [form code] represents the form of the sentence. In the pre-training stage, we incorporate all the pre-training data of five figures of speech (Section~\ref{sec:data}) by concatenating data: $D = \{D_{1}, ..., D_{i}\}$ where each $D_{i}$ is a collection of texts in a figurative form. Following~\citet{liu-etal-2020-multilingual-denoising}, our model is trained on a denoising task, where it is asked to reconstruct text from a version corrupted with a noise function that randomly masks 35\% of the words in the sentence. The [form code] is used as the initial token to predict the sentence (Figure~\ref{fig:overview-bart}~(top)).

\subsection{Literal$\leftrightarrow$Figurative Form Generation}
In \textit{Literal$\leftrightarrow$Figurative} generation, the model generates a text with the desired figure of speech given a literal text, or viceversa.
First, following~\citet{lai-etal-2021-generic}, we use the parallel paraphrase pre-training data to make the model learn the basic task of rewriting. In practice, we incorporate all the data and add the corresponding form code at the beginning of each sentence to train the model in a supervised regime. Second, we fine-tune the model with the literal$\leftrightarrow$figurative parallel data (Table~\ref{tab:data-statistics}) in the same way (PT-to-FT; Figure~\ref{fig:overview-bart} (down)). Since hyperbole and idiom datasets are too small, we upsample them by replication obtaining training sets of 10,000 sentence pairs.

\subsection{Figurative$\leftrightarrow$Figurative Form Generation}
In \textit{Figurative$\leftrightarrow$Figurative} generation, the model takes a text with a given figurative form, and generates a text with the target figurative form. It is important to note that this procedure can have two outcomes: the target figure of speech \textit{substitutes} the original one, or it is \textit{added} to it, yielding a text that contains both the original and the target figurative forms. 

Specifically, given a sentence of tokens $\boldsymbol{x} =\{x_{1}, \cdots, x_{n}\}$ with the figure of speech $s$, the model is asked to generate the corresponding sequence $\boldsymbol{y} =\{y_{1}, \cdots, y_{m}\}$ with the target figure of speech $t$. To overcome the lack of parallel data in different figures of speech which would be necessary to train such a model, we design a mechanism which can leak the information of the desired figure of speech to the encoder with a figurative embedding as additional input. Formally, we employ cross attention to inject the figurative information into word embedding of the input in the fine-tuning process (mFLAG; Figure~\ref{fig:overview-encoder}).

\begin{equation}
\label{eq:cross}
    \textrm{CrossAttn}(\mathbf{W}, \mathbf{F}) = \textrm{softmax}(\frac{\mathbf{W}\mathbf{F}^{\text{T}}}{\sqrt{d}})\mathbf{F}\\
\end{equation}
where $\mathbf{W} \in \mathbb{R}^{m \times d}$ represents the embedding of the source sentence. $\mathbf{F} \in \mathbb{R}^{1 \times d}$ is the embedding of the target form code $T$. To avoid introducing new parameters and catastrophic forgetting, we do not use the commonly used feed-forward block here. We also employ a residual connection~\citep{he-etal-2016} for the word embedding:

\begin{equation}
\label{eq:residual}
    \mathbf{C} = \textrm{CrossAttn}(\mathbf{W}, \mathbf{F}) + \mathbf{W}\\
\end{equation}

\noindent The probability of the output can be computed conditioned both on the input sentence $\boldsymbol{x}$ and the target form code $T$. It can be formulated as:

\begin{equation}
p_{\theta}(\boldsymbol{y} | \boldsymbol{x}, T) 
= \prod_{t=1}^{m} p_{\theta}(y_t | y_{1,...,t-1}; \textrm{\textbf{C}}))
\end{equation}
We also first use the pre-training data to enhance model's rewriting ability, and employ upsampling to augment the gold training data for hyperbole and idiom. 
We use two settings for generation: (i) the model generates text in the target form directly from the source form (mFLAG-DR), meaning that direct figurative-figurative transformation is achieved; (ii) the model uses literal forms as pivot: it first transforms the source text back into its literal form, and then uses this obtained literal form to generate in the target form (mFLAG-BT). Comparing these two models will contribute to better understand the benefits of modelling multi-figurative language generation directly.

\section{Experiments}
All experiments are implemented atop Transformers~\citep{wolf-etal-2020-transformers} using BART-large~\citep{lewis-etal-2020-bart}.  
We train models with batch size 32, accumulating gradients over 8 update steps, using the Adam optimiser~\citep{kingma2017adam} with learning rate 1e-5. We use early stopping (patience 5) if validation performance does not improve. 

\subsection{Evaluation Method}
To assess the model performance we use automatic metrics commonly used in figurative language generation and text style transfer, which focus on form strength and context preservation.

\begin{table}[t]
\resizebox{\linewidth}{!}{%
\centering
\footnotesize
\begin{tabular}{l|ccc}
\toprule
  \textbf{Forms} & \textbf{Precision Score} & \textbf{Recall Score} & \textbf{F1 Score}\\
  \midrule
  Hyperbole & 0.858 & 0.967 & 0.909\\
  Idiom & 0.897 & 0.961 & 0.928\\
  Sarcasm & 0.763 & 0.847 & 0.803\\
  Metaphor & 0.716 & 0.707 & 0.711\\
  Simile & 1.000 & 0.700 & 0.824\\
\bottomrule
\end{tabular}
}
\caption{\label{tab:classifier}
 Accuracy of classifiers for different forms.}
\end{table}

\paragraph{Form Strength} 
To evaluate the form accuracy of the generated text, we  reuse the binary classifiers trained for selecting pre-training data. High confidence for the target figurative form, suggests high accuracy in the generation.  
The performance of the classifiers on the test set (Table~\ref{tab:classifier}), suggests that they are very reliable for Simile, Idiom, and Hyperbole, and slightly less for Metaphor and Sarcasm.

\paragraph{Context Preservation} To assess this aspect, we adopt BLEU and BERTScore (F1-Score)~\citep{bert-score} following previous work~\citep{chakrabarty-etal-2020-generating, zhang2021mover, zhou2021solving, tian-etal-2021-hypogen-hyperbole}. In addition, we employ BLEURT~\citep{sellam-etal-2020-bleurt} and COMET~\citep{rei-etal-2020-comet}, two learnable metrics that have shown promising results in the evaluation of formality transfer~\citep{lai-etal-2022-human}. For all metrics, we calculate scores between model outputs and references for the literal$\leftrightarrow$figurative generation, and between outputs and source sentences (and literal sentences) for figurative$\leftrightarrow$figurative generation as the latter has no parallel data available. \footnote{In our evaluation, we take \texttt{multi-bleu.perl} to calculate BLEU score, and models \texttt{bleurt-large-512} and \texttt{wmt-large-da-estimator-1719} for BLEURT and COMET, respectively.}



\paragraph{Overall Score} We compute the harmonic mean (HM) of figurative accuracy and BLEU score for a direct comparison to baselines.

\begin{table*}[!t]
\setlength{\tabcolsep}{4.5pt}
\resizebox{\linewidth}{!}{%
\centering
\footnotesize
\begin{tabular}{l|cccccc|cccccc}
\toprule
  & \textbf{TGT} & \textbf{BLEU} & \textbf{BERT} & \textbf{BLEURT} & \textbf{COMET} & \textbf{HM} & \textbf{TGT} & \textbf{BLEU} & \textbf{BERT} & \textbf{BLEURT} & \textbf{COMET} & \textbf{HM}\\
  \cline{2-13}
   & \multicolumn{6}{c|}{Literal Form$\rightarrow$Hyperbole} & \multicolumn{6}{c}{Literal Form$\rightarrow$Idiom}\\
  \midrule
  BART-Single & 0.627 & 0.513 & 0.693 & 0.280 & 0.461 & 0.564 & 0.711 & \textbf{0.791} & \textbf{0.855} & \textbf{0.595} & \textbf{0.808} & 0.749\\
  BART-Multi  & 0.707 & 0.541 & 0.698 & 0.260 & 0.352 & 0.613 & 0.637 & 0.747 & 0.829 & 0.498 & 0.706 & 0.688\\
  PT-to-FT    & 0.833 & \textbf{0.582} & \textbf{0.733} & \textbf{0.379} & \textbf{0.490} & \textbf{0.686} & \textbf{0.769} & 0.765 & 0.841 & 0.536 & 0.738 & \textbf{0.767}\\
  mFLAG       & \textbf{0.844} & 0.556 & 0.726 & 0.349 & 0.463 & 0.670 & 0.764 & 0.761 & 0.839 & 0.539 & 0.735 & 0.762\\
  \midrule
   & \multicolumn{6}{c|}{Literal Form$\rightarrow$Sarcasm} & \multicolumn{6}{c}{Literal Form$\rightarrow$Metaphor}\\
  \midrule
  BART-Single & 0.679 & \textbf{0.491} & \textbf{0.611} & \textbf{0.052} & \textbf{0.188} & 0.570 & 0.720 & 0.595 & 0.771 & 0.364 & 0.720 & 0.652\\
  BART-Multi  & 0.743 & 0.483 & 0.598 & 0.011 & 0.137 & 0.585 & 0.767 & 0.577 & 0.780 & 0.434 & 0.785 & 0.659\\
  PT-to-FT    & \textbf{0.765} & 0.485 & 0.609 & 0.040 & 0.162 & \textbf{0.594} & 0.867 & \textbf{0.643} & \textbf{0.812} & \textbf{0.493} & 0.842 & \textbf{0.738}\\
  mFLAG       & 0.762 & 0.487 & 0.609 & 0.043 & 0.169 & \textbf{0.594} & \textbf{0.880} & 0.628 & 0.809 & 0.490 & \textbf{0.844} & 0.733\\
  \midrule
  & \multicolumn{6}{c|}{Literal Form$\rightarrow$Simile} & \multicolumn{6}{c}{Figurative$\rightarrow$Literal Form} \\
  \midrule
  BART-Single & 0.647 & 0.724 & 0.720 & \textbf{0.017} & \textbf{0.321} & 0.683 & 0.733 & 0.606 & 0.742 & 0.284 & 0.455 & 0.663\\
  BART-Multi  & 0.420 & 0.658 & 0.681 &-0.025 & 0.178 & 0.513 & 0.725 & 0.622 & 0.762 & 0.364 & 0.522 & 0.670\\
  PT-to-FT    & 0.907 & 0.729 & 0.722 &-0.021 & 0.219 & 0.808 & \textbf{0.801} & 0.634 & 0.766 & \textbf{0.542} & 0.544 & \textbf{0.708}\\
  mFLAG       & \textbf{0.953} & \textbf{0.745} & \textbf{0.727} &-0.021 & 0.220 & \textbf{0.836} & 0.796 & \textbf{0.637} & \textbf{0.769} & 0.375 & \textbf{0.681} & 0.707\\
\bottomrule
\end{tabular}}
\caption{\label{tab:results-one}
 Results of literal$\leftrightarrow$figurative form generation. TGT represents the accuracy of output labeled as the target form by the classifier; the results of figurative$\rightarrow$literal form generation are averaged across all figures of speech.}
\end{table*}

\begin{table*}[!t]
\setlength{\tabcolsep}{4.5pt}
\resizebox{\linewidth}{!}{%
\centering
\footnotesize
\begin{tabular}{p{1.7cm}|cc|ccccc|ccccc}
\toprule
  & \multicolumn{2}{c|}{\textbf{Form Strength}} & \multicolumn{5}{c|}{\textbf{Source Text}} & \multicolumn{5}{c}{\textbf{Literal Text}} \\
  \cline{2-13}
  & \textbf{SRC} & \textbf{TGT} & \textbf{BLEU} & \textbf{BERT} & \textbf{BLEURT} & \textbf{COMET} & \textbf{HM} &  \textbf{BLEU} & \textbf{BERT} & \textbf{BLEURT} & \textbf{COMET} & \textbf{HM}\\
  \hline
  \multicolumn{13}{c}{Hyperbole$\rightarrow$Others} \\
  \midrule
  BART-Single & 0.470 & 0.425 & 0.665 & 0.782 & 0.459 & 0.472 & 0.519 & 0.488 & 0.700 & 0.294 & 0.248 & 0.454\\
  BART-Multi  & 0.328 & 0.242 & 0.602 & 0.761 & 0.455 & 0.443 & 0.345 & 0.505 & 0.731 & 0.427 & 0.385 & 0.327\\
  PT-to-FT    & 0.252 & 0.258 & 0.590 & 0.749 & 0.437 & 0.420 & 0.359 & \textbf{0.507} & \textbf{0.732} & \textbf{0.438} & \textbf{0.407} & 0.342\\
  mFLAG-DR    & \textbf{0.922} & 0.608 & \textbf{0.815} & \textbf{0.893} & \textbf{0.753} & \textbf{0.836} & \textbf{0.696} & 0.411 & 0.633 & 0.036 &-0.105 & 0.490\\
  mFLAG-BT    & 0.482 & \textbf{0.644} & 0.539 & 0.702 & 0.253 & 0.246 & 0.586 & 0.421 & 0.662 & 0.169 & 0.093 & \textbf{0.509}\\
  \midrule
  \multicolumn{13}{c}{Idiom$\rightarrow$Others}\\
  \midrule
  BART-Single & 0.290 & 0.309 & 0.783 & 0.864 & 0.575 & 0.646 & 0.443 & 0.749 & 0.844 & 0.578 & 0.659 & 0.438\\
  BART-Multi  & 0.273 & 0.204 & 0.785 & 0.873 & 0.602 & 0.674 & 0.324 & 0.758 & 0.859 & 0.630 & 0.701 & 0.408\\
  PT-to-FT    & 0.204 & 0.207 & 0.771 & 0.867 & 0.594 & 0.662 & 0.326 & \textbf{0.760} & \textbf{0.860} & \textbf{0.646} & \textbf{0.715} & 0.325\\
  mFLAG-DR    & \textbf{0.910} & 0.400 & \textbf{0.901} & \textbf{0.940} & \textbf{0.822} & \textbf{0.869} & \textbf{0.554} & 0.694 & 0.799 & 0.328 & 0.375 & 0.507\\
  mFLAG-BT    & 0.328 & \textbf{0.409} & 0.724 & 0.831 & 0.491 & 0.566 & 0.523 & 0.703 & 0.816 & 0.490 & 0.569 & \textbf{0.517}\\
  \midrule
  \multicolumn{13}{c}{Sarcasm$\rightarrow$Others} \\
  \midrule
  BART-Single & 0.577 & 0.370 & 0.877 & 0.899 & 0.650 & 0.792 & 0.520 & 0.454 & 0.579 &-0.088 &-0.051 & 0.408\\
  BART-Multi  & 0.569 & 0.247 & 0.903 & 0.923 & 0.701 & 0.838 & 0.388 & 0.471 & \textbf{0.593} &-0.049 &-0.014 & 0.324\\
  PT-to-FT    & 0.464 & 0.252 & 0.863 & 0.891 & 0.613 & 0.774 & 0.390 & \textbf{0.468} & 0.592 &\textbf{-0.031} & \textbf{0.000} & 0.328\\
  mFLAG-DR    & \textbf{0.840} & 0.438 & \textbf{0.907} & \textbf{0.928} & \textbf{0.813} & \textbf{0.872} & 0.591 & 0.442 & 0.563 &-0.198 &-0.143 & 0.440\\
  mFLAG-BT    & 0.583 & \textbf{0.481} & 0.808 & 0.831 & 0.460 & 0.604 & \textbf{0.605} & 0.430 & 0.554 &-0.164 &-0.133 & 0.\textbf{454}\\
  \midrule
  \multicolumn{13}{c}{Metaphor$\rightarrow$Others} \\
  \midrule
  BART-Single & 0.163 & 0.314 & 0.603 & 0.776 & 0.412 & 0.555 & 0.413 & 0.575 & 0.773 & 0.381 & 0.486 & 0.406\\
  BART-Multi  & 0.255 & 0.249 & 0.647 & 0.825 & 0.554 & 0.723 & 0.360 & 0.632 & 0.820 & 0.550 & 0.689 & 0.357\\
  PT-to-FT    & 0.147 & 0.254 & 0.671 & 0.832 & 0.599 & \textbf{0.763} & 0.369 & \textbf{0.648} & \textbf{0.824} & \textbf{0.507} & \textbf{0.665} & 0.365\\
  mFLAG-DR    & \textbf{0.795} & 0.518 & \textbf{0.697} & \textbf{0.846} & \textbf{0.614} & 0.706 & \textbf{0.594} & 0.516 & 0.758 & 0.320 & 0.410 & 0.517\\
  mFLAG-BT    & 0.387 & \textbf{0.557} & 0.502 & 0.734 & 0.329 & 0.434 & 0.528 & 0.496 & 0.743 & 0.317 & 0.417 & \textbf{0.525}\\
  \midrule
  \multicolumn{13}{c}{Simile$\rightarrow$Others}\\
  \midrule
  BART-Single & 0.057 & 0.607 & 0.469 & 0.559 &-0.406 &-0.429 & 0.529 & 0.588 & 0.667 & 0.160 &-0.102 & 0.597\\
  BART-Multi  & 0.007 & 0.272 & 0.629 & 0.686 &-0.043 &-0.051 & 0.380 & \textbf{0.765} & \textbf{0.818} & \textbf{0.262} & \textbf{0.415} & 0.401\\
  PT-to-FT    & 0.000 & 0.314 & 0.622 & 0.671 &-0.031 &-0.067 & 0.417 & 0.754 & 0.804 & 0.244 & 0.394 & 0.443\\
  mFLAG-DR    & \textbf{0.440} & 0.685 & \textbf{0.849} & \textbf{0.884} & \textbf{0.637} & \textbf{0.690} & \textbf{0.758} & 0.589 & 0.698 &-0.016 &-0.057 & 0.633\\
  mFLAG-BT    & 0.132 & \textbf{0.687} & 0.606 & 0.670 &-0.069 &-0.064 & 0.644 & 0.672 & 0.766 & 0.163 & 0.250 & \textbf{0.679}\\
\bottomrule
\end{tabular}}
\caption{\label{tab:results-many}
 Results of figurative$\leftrightarrow$figurative form generation. Notes: (i) SRC (TGT) represents the accuracy of output labeled as the source (target) form by the classifier of the source (target) form; (ii) results for each block are averaged for all generations from one figurative language to others.}
\end{table*}

\subsection{Baselines}
We compare our systems to two strong baselines.

\paragraph{BART-Single} For each figure of speech, we fine-tune BART on the corresponding parallel data. For figurative$\rightarrow$figurative generation, we  use each figurative-to-literal model to generate the literal text, and then feed it into the model of the target form to generate the output.

\paragraph{BART-Multi} We concatenate the five parallel training sets and fine-tune BART for multi-figurative language modelling, thereby enabling the generation between different forms.

\subsection{Literal$\leftrightarrow$Figurative Generation}
Table~\ref{tab:results-one} presents the results of literal$\leftrightarrow$figurative form generation. BART-Multi outperforms BART-Single on most generation directions, except literal-to-idiom and literal-to-simile. This suggests that the model does benefit from multi-figurative language modelling with cross-figurative knowledge transfer. Compared to BART-Single and BART-Multi, both of our proposed models PT-to-FT and mFLAG have consistently stronger results. Specifically, we observe that BART-Single has the best performance only on context preservation for literal-to-idiom and literal-to-sarcasm generation, while our models are better for the rest, especially with a good balance between form strength and context preservation. The results confirm that our pre-training scheme and strategies significantly improve performances for multi-figurative language modelling. When looking at PT-to-FT and mFLAG, we see that these two models' performances are very close on all tasks and do not show a clear and consistent trend. The main reason for this is most likely that the settings of the two models are almost identical except that mFLAG has a figurative injection mechanism, and they are both trained with parallel literal$\leftrightarrow$figurative sentence pairs. 

\begin{table*}[!t]
\footnotesize
\resizebox{\linewidth}{!}{%
\centering\begin{tabular}{l|l|p{10cm}|c}
\toprule
 \textbf{Forms} & \textbf{Models} & \textbf{Sentences} & \textbf{Suc.}\\
 \midrule
 Hyperbole & Input  & I am not happy that he urged me to finish \textcolor{red}{all the hardest tasks in the world}. & -\\
 \midrule
 \multirow{4}{*}{Literal} & BART-Single & I am not happy that he urged me to finish all the hard tasks in the world. & \XSolidBrush\\
 & BART-Multi & I am glad that he urged me to finish all the hardest tasks in the world. & \XSolidBrush\\
 & PT-to-FT   & I am not happy that he urged me to complete all the difficult tasks in the world. & \XSolidBrush\\
 & mFLAG   & I am not happy that he urged me to finish \textcolor{red}{all the difficult tasks}. & \textcolor{red}{\CheckmarkBold}\\
 \midrule
 \multirow{5}{*}{Idiom} & BART-Single & I am not happy that he urged me to finish all the hard tasks in the world. & \XSolidBrush\\
 & BART-Multi & I am so glad that he urged me to finish all the hardest tasks in the world. & \XSolidBrush\\
 & PT-to-FT   & I am not happy that he urged me to complete all the difficult tasks. & \XSolidBrush\\
 & mFLAG-DR   & I am not happy that he urged me to \textcolor{red}{tick off} all the hardest tasks in the world. & \textcolor{red}{\CheckmarkBold}\\
 & mFLAG-BT   & I am not happy that he urged me to \textcolor{red}{get cracking} on all the difficult tasks. & \textcolor{red}{\CheckmarkBold}\\
 \midrule
 \multirow{5}{*}{Sarcasm} & BART-Single& I'm \textcolor{red}{so glad} my dad encouraged me to finish all the hard tasks in the world. & \textcolor{red}{\CheckmarkBold}\\
 & BART-Multi & \textcolor{red}{Thank you for encouraging me} to finish all the hardest tasks in the world. & \textcolor{red}{\CheckmarkBold}\\
 & PT-to-FT   & I am \textcolor{red}{happy} that he urged me to finish all the hardest tasks in the world. & \textcolor{red}{\CheckmarkBold}\\
 & mFLAG-DR   & \textcolor{red}{Glad} he urged me to finish all the hardest tasks in the world. & \textcolor{red}{\CheckmarkBold}\\
 & mFLAG-BT   & \textcolor{red}{Glad} he urged me to finish all the difficult tasks. & \textcolor{red}{\CheckmarkBold}\\
\midrule
 \multirow{5}{*}{Metaphor} & BART-Single & I am not happy that he urged me to \textcolor{red}{bear} all the difficult tasks.  & \textcolor{red}{\CheckmarkBold}\\
 & BART-Multi & I am so glad that he urged me to finish all the hardest tasks in the world. & \XSolidBrush\\
 & PT-to-FT   & I am not happy that he urged me to complete all the difficult tasks in the world. & \XSolidBrush\\
 & mFLAG-DR   & I am not happy that he urged me to \textcolor{red}{bear} all the hardest tasks in the world.  & \textcolor{red}{\CheckmarkBold}\\
 & mFLAG-BT & I am not happy that he pressed me to finish all the difficult tasks. & \XSolidBrush\\
 \midrule
 \multirow{5}{*}{Simile} & BART-Single  & I am not happy that he urged me to finish all the difficult tasks. & \XSolidBrush\\
 & BART-Multi   & I am so glad that he urged me to finish all the hardest tasks in the world. & \XSolidBrush\\
 & PT-to-FT     & I am not happy that he urged me to complete all the difficult tasks in the world. & \XSolidBrush\\
 & mFLAG-DR    & I am not happy that he urged me to finish all the \textcolor{red}{like a million things}. & \textcolor{red}{\CheckmarkBold}\\
 & mFLAG-BT & I am not happy that he urged me to finish all the difficult tasks. & \XSolidBrush\\
\bottomrule
\end{tabular}}
\caption{\label{tab:cases}
 Examples outputs generated by various models from hyperbolic text, where \textcolor{red}{red} denotes appropriate words/phrases for desired forms. Suc.==Successful.}
\end{table*}

\subsection{Figurative$\leftrightarrow$Figurative Generation}
Table~\ref{tab:results-many} reports results of figurative$\leftrightarrow$figurative form generation.\footnote{Complete results are in Appendix~\ref{app:fig-to-fig}.} We see that both BART-Multi and PT-to-FT perform poorly on the form strength and the context preservation computed against the source text. 
The low form strength (SRC and TGT, see table's caption) and high scores of context preservation (using literal text) suggest that these two models transform the source text into the literal form.
BART-Single, interestingly, shows a better performance on both form strength and context preservation. For mFLAG-DR and mFLAG-BT, we see that they show the best performance across the board: (i)  mFLAG-DR shows a significant improvement in target figurative form (TGT) while maintaining the original form (SRC) very much; it also achieves the best performance on context preservation; 
(ii) mFLAG-BT achieves the highest form accuracy in the target figure of speech while reducing the original form strength. 

It is interesting to note that the direct generation method might allow for the source figure of speech to be retained in the generated sentence, as we do not explicitly remove it by transforming the sentence to its literal form first. For example, with hyperbolic input ``\textit{I am not happy that he urged me to finish all the hard task in the world}'', one of our sarcastic transformations reads ``\textit{Thank you for encouraging me to finish all the hardest tasks in the world}", where the hyperbolic part (``\textit{all the hardest tasks in the world}") is preserved unchanged (see Table~\ref{tab:cases}). 

Overall, the results show that mFLAG with the mechanism for injecting the figurative information into the encoder can generate from one figure of speech to another even without task-specific parallel data.

\section{Analysis and Discussion}

\paragraph{Case Study}
Table~\ref{tab:cases} shows a group of example outputs for hyperbole$\rightarrow$others generated by various models.\footnote{More example outputs of mFLAG are in Appendix~\ref{app:outputs}.} From the results of hyperbole$\rightarrow$literal generation, we see that mFLAG generates the literal sentence from the hyperbolic one very well, confirming that texts generated by mFLAG-BT tend to contain less the source form by substituting it with the target form. In figurative$\leftrightarrow$figurative generation, all models nicely generate sarcastic text while all baselines usually fail at generating the other forms. Since the metaphor generation dataset we used focuses on metaphorical verb aspect, we consider the outputs of BART-Single and mFLAG-DR to be successful. Overall, our proposed mFLAG based models perform better on all generation directions.

\begin{figure}[!t]
    \centering
    \subfigure[Result for PT-to-FT.]{
      \includegraphics[scale=0.45]{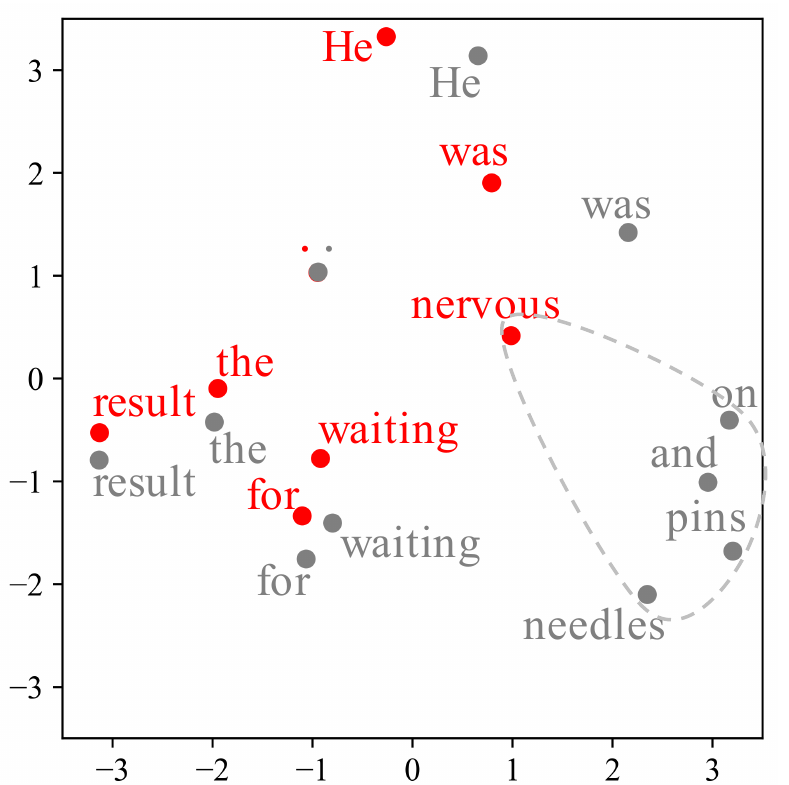}
      \label{fig:pca-ft-literal-idiom}
    }
    \subfigure[Result for mFLAG.]{
        \includegraphics[scale=0.44]{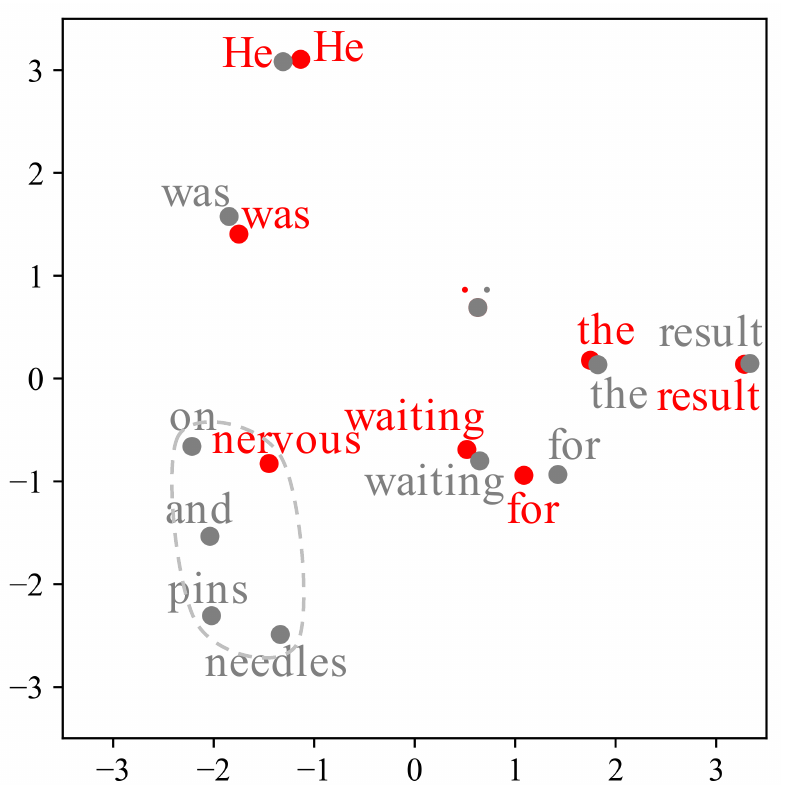} 
        \label{fig:pca-ca-literal-idiom}
    }
    
    \subfigure[Result for PT-to-FT.]{
      \includegraphics[scale=0.45]{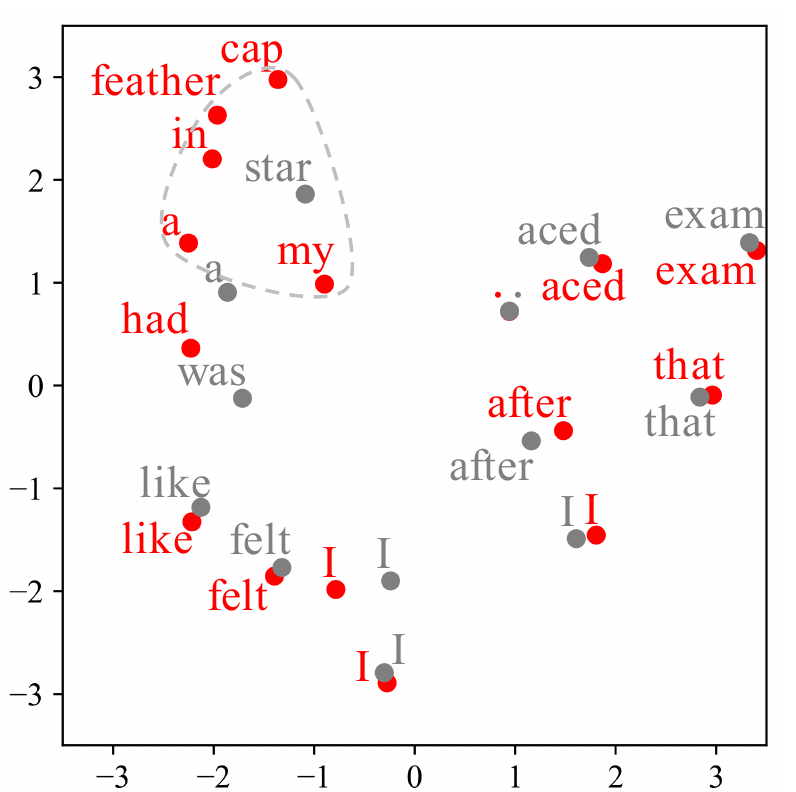}
      \label{fig:pca-ft-idiom-hyperbole}
    }
    \subfigure[Result for mFLAG.]{
        \includegraphics[scale=0.44]{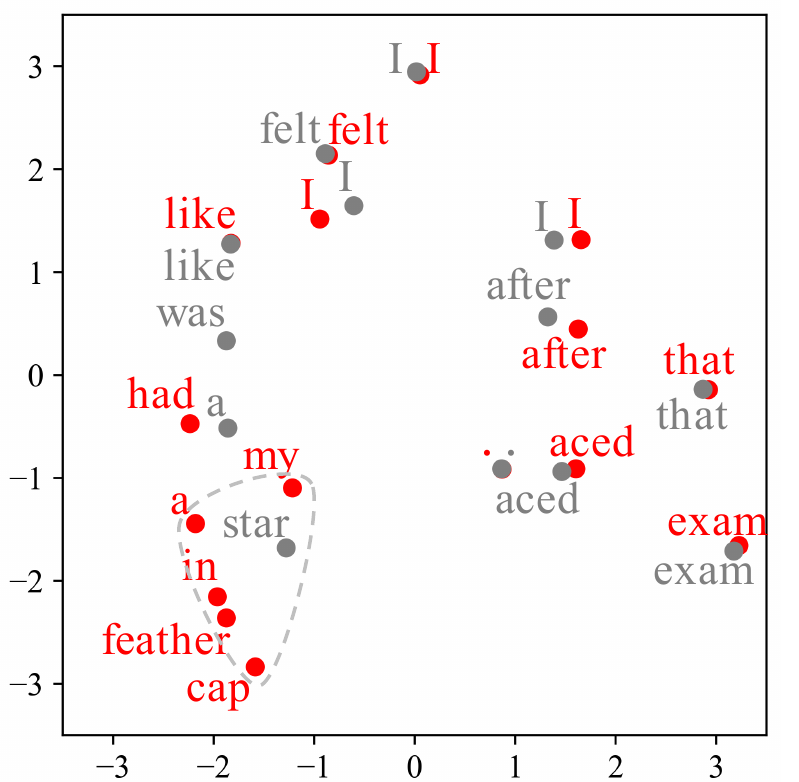} 
        \label{fig:pca-ca-idiom-hyperbole}
    }
    \caption{PCA token representations of encoder outputs for literal$\rightarrow$hyperbole (top) and idiom$\rightarrow$hyperbole (down).} 
    \label{fig:pca-representation}
\end{figure}

\paragraph{Probing Figurative Information for Encoder}
To measure the distribution of source and target sentences encoded by the Encoder with/without the mechanism of injecting figurative information, we apply Principal Component Analysis (PCA) to reduce the dimensionality of the Encoder outputs and visualise relations between tokens in a two-dimensional space. 
Fig.~\ref{fig:pca-ft-literal-idiom} and~\ref{fig:pca-ca-literal-idiom} show the results of a source literal text ``\textit{He was nervous waiting for the result.}'' and a target hyperbolic text ``\textit{He was on pins and needles waiting for the result.}''.
We see the word ``He'' and 'was'' of the two sentences are not in the same cluster in ~\ref{fig:pca-ft-literal-idiom} while it is interesting to see that all distances between token pairs of~\ref{fig:pca-ca-literal-idiom} are closer, especially the phrase ``on pins and needles'', and ``nervous'' are almost in the same cluster in~\ref{fig:pca-ca-literal-idiom}. Fig.~\ref{fig:pca-ft-idiom-hyperbole} and~\ref{fig:pca-ca-idiom-hyperbole} show the results of a source idiomatic text ``\textit{I felt like I had a feather in my cap after I aced that exam.}'' and a target hyperbolic text ``\textit{I felt like I was a star after I aced that exam.}''.
We observe that the token pairs like ``I'', ``like'' and 'felt'' of mFLAG are closer than those of PT-to-FT. It is also interesting to see that the phrase ``a feather in my cap'' and the token ``star'' make more of a cluster in~\ref{fig:pca-ca-idiom-hyperbole}. We believe this benefits the decoder, especially decoding into the target figurative form.

\begin{figure}[t]
    \centering
    \includegraphics[scale=0.56]{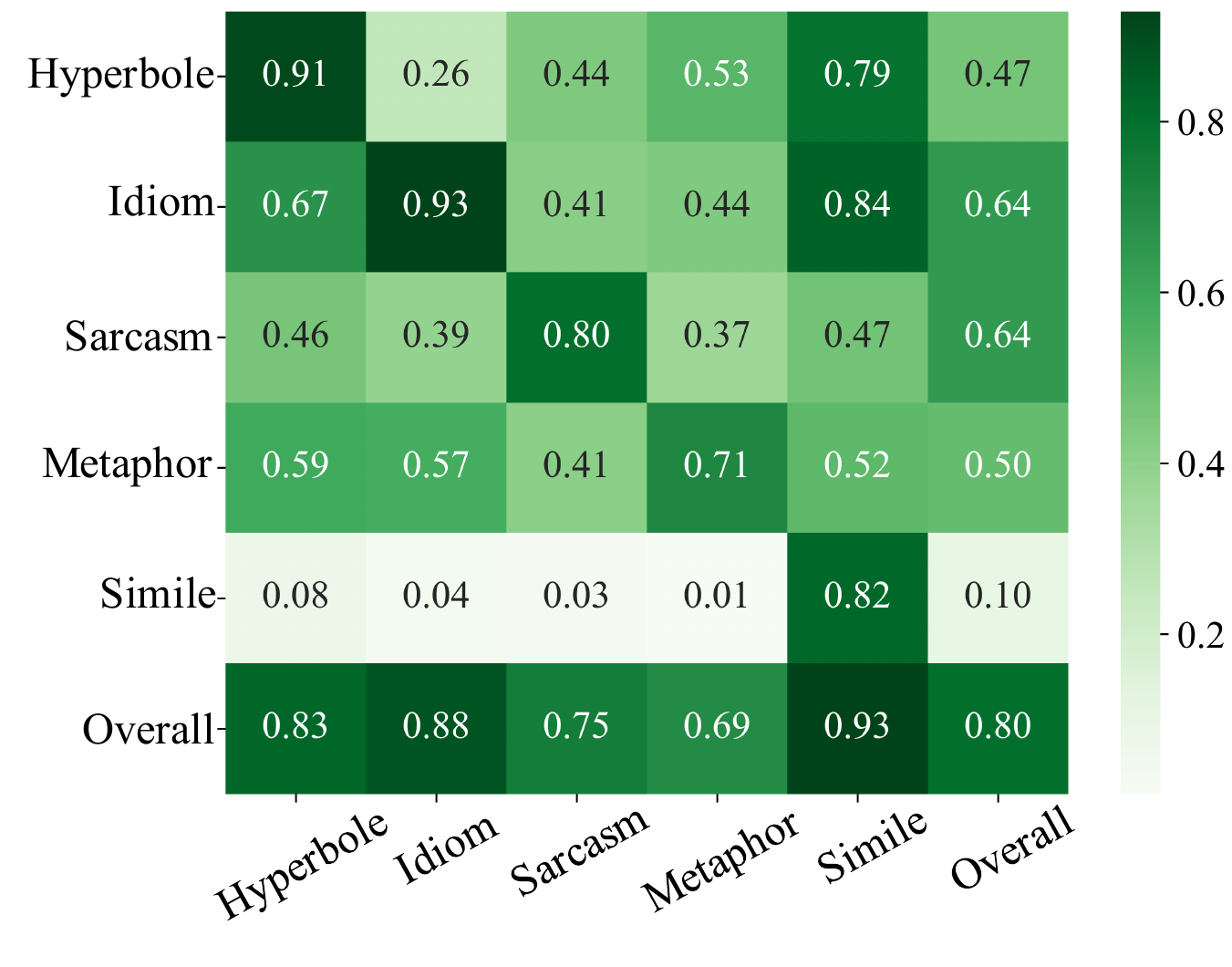}
    \caption{Performances (F1 score) of classifiers on different figurative forms. Each row represents results of a classifier tested on each/all figurative form(s).}
    \label{fig:classification}
\end{figure}

\paragraph{How similar are different forms?} To analyze the connection between literal and figurative forms, and between different figures of speech, we evaluate each figurative classifier on the test sets of the other figurative forms (Figure~\ref{fig:classification}). 
We first see that the overall model (literal vs figurative) achieves F1 scores of over 0.69 for each figure of speech, confirming the feasibility of multi-figurative modelling. For each figure of speech, we observe: (i) classifiers for hyperbole and idiom have high F1 scores on the test set of simile (0.79 and 0.84), suggesting that sentences with similes may also be hyperbolic or idiomatic; (ii) for sarcasm and metaphor, classifiers have medium scores on other forms; (iii) the classifier of simile achieves F1 scores of less than 0.11 on other figures of speech; this is due to the fact that the simile dataset was created using the format \textit{like a}, which is easy for the model to learn. Different figurative forms are related to each other, confirming that models can benefit from cross-figurative knowledge transfer. Further (computational) analysis of similarities and differences will help to even better leverage such transfer.

\section{Conclusion and Outlook}
We have proposed a novel task of multi-figurative language generation, and shown that our models do benefit from cross-figurative knowledge transfer. Paraphrasing data can be leveraged in further pre-training to enhance both form strength and context preservation in figurative language generation. We have also proposed a mechanism for injecting the target figurative information into the encoder, so that we can achieve generation between different figures of speech even without parallel figurative-figurative pairs.

While we innovatively explore multi-figurative language  generation across literal and five figurative forms, and our model achieves the best performances compared to baselines, there is still substantial room for improvement and further extensions.

The current lack of human references for automating the evaluation of figurative-to-figurative generation is surely a limitation in terms of better understanding of the models' behaviour and potential improvements.
More in general, figurative language generation is a relatively new task, which still lacks standardised evaluation methods, both in terms of automatic metrics and human-based evaluation.

Also, we introduce for the first time generation across literal expressions and five figurative forms, but there are many more forms of creative writing that could be modelled. Moreover, we only limited our attention to English, due to data availability, but are convinced that datasets in other languages would greatly benefit research in this area. Indeed, multilingual modelling would make it possible to make connections across different languages, thus shedding more light on cross-lingual regularities in figurative language use, and thus also open up potential avenues to tackle this task better. 

\section*{Acknowledgments}

This work was partly funded by the China Scholarship Council (CSC). The COLING anonymous reviewers provided us with useful comments which contributed to improving this paper and its presentation, so we’re grateful to them. We would also like to thank the Center for Information Technology of the University of Groningen for their support and for providing access to the Peregrine high performance computing cluster.

\bibliography{anthology,custom}
\bibliographystyle{acl_natbib}
\onecolumn
\clearpage
\appendix


\section{\Large Appendices: \\~ \\ }
\label{sec:appendix}

\setcounter{table}{0}
\renewcommand{\thetable}{A.\arabic{table}}
\setcounter{figure}{0}
\renewcommand{\thefigure}{A.\arabic{figure}}

This appendices include: (i) Dataset statistics of pre-training data (\ref{app:pre-training-data}); (ii) Detailed results for figurative$\leftrightarrow$figurative generation (\ref{app:fig-to-fig}); (iii) Example outputs of mFLAG (\ref{app:outputs}) \vspace*{0.5cm}.

\subsection{Pre-Training Data}
\label{app:pre-training-data}
\begin{table}[!ht]
\centering
\footnotesize
\begin{tabular}{llccc}
\toprule[1pt]
 \textbf{Forms} & \textbf{Task} & $\sigma$ & \textbf{Train} & \textbf{Valid}\\
 \hline
 Hyperbole & Literal text$\leftrightarrow$Hyperbole & 0.94 & 102,887 & 5,000\\
 Idiom     & Literal text$\leftrightarrow$idiom     & 0.95 & 133,285 & 5,000\\
 Sarcasm     & Literal text$\leftrightarrow$Sarcasm     & 0.70 & 22,550 & 5,000\\
 Metaphor  & Literal text$\leftrightarrow$Metaphor  & 0.95 & 206,554 & 5,000\\
 Simile    & Literal text$\leftrightarrow$Simile & 0.76 & 57,566 & 5,000\\
\bottomrule[1pt]
\end{tabular}
\caption{\label{tab:pre-training-data}
Dataset statistics for generic pre-training data. Note that $\sigma$ is the threshold used to select sentence pairs.
}
\end{table}

\subsection{Detailed Results for Figurative$\leftrightarrow$Figurative Generation}
\label{app:fig-to-fig}

\begin{table*}[!ht]
\resizebox{\linewidth}{!}{%
\centering
\footnotesize
\begin{tabular}{l|cc|ccccc|ccccc}
\toprule
  & \multicolumn{2}{c|}{\textbf{Form Strength}} & \multicolumn{5}{c|}{\textbf{Source Text}} & \multicolumn{5}{c}{\textbf{Literal Text}} \\
  \cline{2-13}
  & \textbf{SRC} & \textbf{TGT} & \textbf{BLEU} & \textbf{BERT} & \textbf{BLEURT} & \textbf{COMET} & \textbf{HM} & \textbf{BLEU} & \textbf{BERT} & \textbf{BLEURT} & \textbf{COMET} & \textbf{HM}\\
  \midrule
   \multicolumn{13}{c}{Hyperbole$\rightarrow$Idiom}\\
  \hline
  BART-Single & 0.513 & 0.513 & 0.653 & 0.781 & 0.469 & 0.466 & 0.575 & 0.471 & 0.692 & 0.294 & 0.240 & 0.491\\
  BART-Multi  & 0.313 & 0.233 & 0.595 & 0.755 & 0.439 & 0.425 & 0.335 & 0.505 & 0.730 & 0.429 & 0.385 & 0.386\\
  PT-to-FT    & 0.240 & 0.200 & 0.587 & 0.747 & 0.445 & 0.422 & 0.298 & 0.506 & 0.729 & 0.442 & 0.402 & 0.287\\
  mFLAG-DR    & 0.900 & 0.733 & 0.766 & 0.876 & 0.729 & 0.758 & 0.749 & 0.401 & 0.637 & 0.063 &-0.089 & 0.518\\
  mFLAG-BT    & 0.653 & 0.707 & 0.599 & 0.743 & 0.368 & 0.380 & 0.649 & 0.409 & 0.650 & 0.136 &-0.011 & 0.518\\
  \midrule
  \multicolumn{13}{c}{Hyperbole$\rightarrow$Sarcasm}\\
  \hline
  BART-Single & 0.407 & 0.387 & 0.673 & 0.785 & 0.464 & 0.595 & 0.491 & 0.499 & 0.710 & 0.300 & 0.298 & 0.436\\
  BART-Multi  & 0.333 & 0.313 & 0.601 & 0.760 & 0.464 & 0.447 & 0.412 & 0.500 & 0.730 & 0.427 & 0.386 & 0.385\\
  PT-to-FT    & 0.267 & 0.373 & 0.587 & 0.744 & 0.400 & 0.399 & 0.456 & 0.505 & 0.728 & 0.392 & 0.385 & 0.429\\
  mFLAG-DR    & 0.900 & 0.447 & 0.873 & 0.922 & 0.883 & 0.947 & 0.591 & 0.431 & 0.645 & 0.073 &-0.006 & 0.439\\
  mFLAG-BT    & 0.373 & 0.507 & 0.545 & 0.699 & 0.283 & 0.265 & 0.525 & 0.442 & 0.678 & 0.233 & 0.233 & 0.472\\
  \midrule
  \multicolumn{13}{c}{Hyperbole$\rightarrow$Metaphor}\\
  \hline
  BART-Single & 0.407 & 0.533 & 0.653 & 0.784 & 0.501 & 0.509 & 0.587 & 0.499 & 0.712 & 0.369 & 0.331 & 0.515\\
  BART-Multi  & 0.320 & 0.407 & 0.597 & 0.758 & 0.439 & 0.432 & 0.484 & 0.505 & 0.730 & 0.422 & 0.383 & 0.451\\
  PT-to-FT    & 0.253 & 0.447 & 0.592 & 0.756 & 0.450 & 0.432 & 0.509 & 0.513 & 0.736 & 0.451 & 0.423 & 0.478\\
  mFLAG-DR    & 0.927 & 0.773 & 0.823 & 0.902 & 0.762 & 0.870 & 0.797 & 0.412 & 0.634 & 0.033 &-0.081 & 0.538\\
  mFLAG-BT    & 0.300 & 0.753 & 0.495 & 0.692 & 0.227 & 0.235 & 0.597 & 0.433 & 0.686 & 0.252 & 0.226 & 0.550\\
  \midrule
  \multicolumn{13}{c}{Hyperbole$\rightarrow$Simile}\\
  \hline
  BART-Single & 0.553 & 0.267 & 0.680 & 0.779 & 0.402 & 0.416 & 0.383 & 0.481 & 0.687 & 0.214 & 0.123 & 0.342\\
  BART-Multi  & 0.347 & 0.013 & 0.616 & 0.771 & 0.476 & 0.467 & 0.025 & 0.511 & 0.733 & 0.431 & 0.387 & 0.025\\
  PT-to-FT    & 0.247 & 0.013 & 0.595 & 0.747 & 0.451 & 0.424 & 0.025 & 0.505 & 0.732 & 0.465 & 0.418 & 0.332\\
  mFLAG-DR    & 0.960 & 0.480 & 0.798 & 0.873 & 0.639 & 0.709 & 0.599 & 0.400 & 0.616 &-0.026 &-0.242 & 0.436\\
  mFLAG-BT    & 0.600 & 0.607 & 0.525 & 0.674 & 0.135 & 0.105 & 0.551 & 0.401 & 0.634 & 0.055 &-0.077 & 0.563\\
\bottomrule
\end{tabular}}
\caption{\label{tab:results-hyperbole}
 Results of hyperbole$\rightarrow$others generation. }
\end{table*}

\clearpage

\begin{table*}[!t]
\resizebox{\linewidth}{!}{%
\centering
\footnotesize
\begin{tabular}{l|cc|ccccc|ccccc}
\toprule
  & \multicolumn{2}{c|}{\textbf{Form Strength}} & \multicolumn{5}{c|}{\textbf{Source Text}} & \multicolumn{5}{c}{\textbf{Literal Text}} \\
  \cline{2-13}
  & \textbf{SRC} & \textbf{TGT} & \textbf{BLEU} & \textbf{BERT} & \textbf{BLEURT} & \textbf{COMET} & \textbf{HM} & \textbf{BLEU} & \textbf{BERT} & \textbf{BLEURT} & \textbf{COMET} & \textbf{HM}\\
  \midrule
   \multicolumn{13}{c}{Idiom$\rightarrow$Hyperbole}\\
  \hline
  BART-Single & 0.311 & 0.103 & 0.788 & 0.867 & 0.585 & 0.653 & 0.182 & 0.751 & 0.844 & 0.575 & 0.651 & 0.181\\
  BART-Multi  & 0.269 & 0.031 & 0.784 & 0.872 & 0.600 & 0.671 & 0.059 & 0.758 & 0.859 & 0.632 & 0.702 & 0.059\\
  PT-to-FT    & 0.232 & 0.041 & 0.782 & 0.874 & 0.614 & 0.681 & 0.078 & 0.763 & 0.862 & 0.647 & 0.717 & 0.078\\
  mFLAG-DR    & 0.929 & 0.232 & 0.847 & 0.908 & 0.716 & 0.769 & 0.364 & 0.667 & 0.783 & 0.286 & 0.317 & 0.344\\
  mFLAG-BT    & 0.564 & 0.172 & 0.728 & 0.836 & 0.523 & 0.574 & 0.278 & 0.679 & 0.799 & 0.415 & 0.477 & 0.274\\
  \midrule
  \multicolumn{13}{c}{Idiom$\rightarrow$Sarcasm}\\
  \hline
  BART-Single & 0.277 & 0.335 & 0.795 & 0.872 & 0.602 & 0.671 & 0.471 & 0.761 & 0.853 & 0.609 & 0.692 & 0.465\\
  BART-Multi  & 0.281 & 0.292 & 0.785 & 0.875 & 0.608 & 0.679 & 0.426 & 0.756 & 0.857 & 0.623 & 0.693 & 0.421\\
  PT-to-FT    & 0.230 & 0.319 & 0.773 & 0.866 & 0.587 & 0.657 & 0.452 & 0.755 & 0.854 & 0.620 & 0.690 & 0.449\\
  mFLAG-DR    & 0.924 & 0.376 & 0.927 & 0.955 & 0.871 & 0.919 & 0.535 & 0.711 & 0.804 & 0.345 & 0.395 & 0.492\\
  mFLAG-BT    & 0.233 & 0.405 & 0.721 & 0.828 & 0.485 & 0.570 & 0.519 & 0.710 & 0.821 & 0.515 & 0.613 & 0.516\\
  \midrule
   \multicolumn{13}{c}{Idiom$\rightarrow$Metaphor}\\
  \hline
  BART-Single & 0.280 & 0.692 & 0.768 & 0.858 & 0.561 & 0.643 & 0.728 & 0.734 & 0.840 & 0.571 & 0.667 & 0.728\\
  BART-Multi  & 0.268 & 0.485 & 0.784 & 0.872 & 0.600 & 0.671 & 0.599 & 0.759 & 0.859 & 0.633 & 0.703 & 0.592\\
  PT-to-FT    & 0.170 & 0.467 & 0.762 & 0.862 & 0.581 & 0.656 & 0.579 & 0.760 & 0.862 & 0.656 & 0.728 & 0.579\\
  mFLAG-DR    & 0.866 & 0.798 & 0.879 & 0.938 & 0.821 & 0.876 & 0.837 & 0.687 & 0.803 & 0.359 & 0.420 & 0.739\\
  mFLAG-BT    & 0.247 & 0.798 & 0.703 & 0.828 & 0.482 & 0.580 & 0.747 & 0.688 & 0.820 & 0.515 & 0.620 & 0.739\\
  \midrule
   \multicolumn{13}{c}{Idiom$\rightarrow$Simile}\\
  \hline
  BART-Single & 0.293 & 0.106 & 0.782 & 0.859 & 0.550 & 0.616 & 0.187 & 0.748 & 0.839 & 0.557 & 0.627 & 0.186\\
  BART-Multi  & 0.274 & 0.007 & 0.786 & 0.874 & 0.601 & 0.673 & 0.014 & 0.759 & 0.860 & 0.633 & 0.704 & 0.014\\
  PT-to-FT    & 0.184 & 0.000 & 0.766 & 0.864 & 0.592 & 0.655 & 0.000 & 0.762 & 0.862 & 0.662 & 0.726 & 0.000\\
  mFLAG-DR    & 0.920 & 0.193 & 0.949 & 0.959 & 0.878 & 0.909 & 0.321 & 0.712 & 0.805 & 0.322 & 0.368 & 0.304\\
  mFLAG-BT    & 0.266 & 0.259 & 0.744 & 0.832 & 0.475 & 0.539 & 0.384 & 0.736 & 0.825 & 0.514 & 0.566 & 0.383\\
\bottomrule
\end{tabular}}
\caption{\label{tab:results-idiom}
 Results of idiom$\rightarrow$others generation. }
\end{table*}

\begin{table*}[!ht]
\resizebox{\linewidth}{!}{%
\centering
\footnotesize
\begin{tabular}{l|cc|ccccc|ccccc}
\toprule
  & \multicolumn{2}{c|}{\textbf{Form Strength}} & \multicolumn{5}{c|}{\textbf{Source Text}} & \multicolumn{5}{c}{\textbf{Literal Text}} \\
  \cline{2-13}
  & \textbf{SRC} & \textbf{TGT} & \textbf{BLEU} & \textbf{BERT} & \textbf{BLEURT} & \textbf{COMET} & \textbf{HM} & \textbf{BLEU} & \textbf{BERT} & \textbf{BLEURT} & \textbf{COMET} & \textbf{HM}\\
  \midrule
  \multicolumn{13}{c}{Sarcasm$\rightarrow$Hyperbole}\\
  \hline
  BART-Single & 0.568 & 0.405 & 0.907 & 0.921 & 0.727 & 0.855 & 0.560 & 0.470 & 0.590 &-0.050 &-0.010 & 0.435\\
  BART-Multi  & 0.558 & 0.347 & 0.898 & 0.918 & 0.690 & 0.828 & 0.501 & 0.471 & 0.592 &-0.048 &-0.013 & 0.400\\
  PT-to-FT    & 0.459 & 0.384 & 0.878 & 0.901 & 0.635 & 0.799 & 0.534 & 0.473 & 0.595 &-0.022 & 0.010 & 0.349\\
  mFLAG-DR    & 0.823 & 0.466 & 0.914 & 0.936 & 0.862 & 0.904 & 0.617 & 0.449 & 0.569 &-0.169 &-0.114 & 0.457\\
  mFLAG-BT    & 0.612 & 0.473 & 0.821 & 0.849 & 0.548 & 0.675 & 0.595 & 0.438 & 0.562 &-0.123 &-0.095 & 0.455\\
   \midrule
  \multicolumn{13}{c}{Sarcasm$\rightarrow$Idiom}\\
  \hline
  BART-Single & 0.582 & 0.429 & 0.853 & 0.889 & 0.615 & 0.730 & 0.571 & 0.441 & 0.575 &-0.098 &-0.090 & 0.435\\
  BART-Multi  & 0.568 & 0.299 & 0.901 & 0.921 & 0.697 & 0.836 & 0.449 & 0.472 & 0.593 &-0.051 &-0.017 & 0.366\\
  PT-to-FT    & 0.422 & 0.276 & 0.862 & 0.886 & 0.599 & 0.700 & 0.418 & 0.462 & 0.594 &-0.024 & 0.006 & 0.394\\
  mFLAG-DR    & 0.847 & 0.517 & 0.875 & 0.911 & 0.749 & 0.808 & 0.650 & 0.426 & 0.554 &-0.229 &-0.193 & 0.467\\
  mFLAG-BT    & 0.599 & 0.527 & 0.791 & 0.825 & 0.442 & 0.570 & 0.633 & 0.417 & 0.550 &-0.176 &-0.166 & 0.466\\
  \midrule
  \multicolumn{13}{c}{Sarcasm$\rightarrow$Metaphor}\\
  \hline
  BART-Single & 0.571 & 0.483 & 0.851 & 0.881 & 0.591 & 0.788 & 0.616 & 0.445 & 0.571 &-0.112 &-0.049 & 0.463\\
  BART-Multi  & 0.561 & 0.337 & 0.900 & 0.919 & 0.693 & 0.830 & 0.490 & 0.471 & 0.592 &-0.046 &-0.014 & 0.393\\
  PT-to-FT    & 0.514 & 0.344 & 0.870 & 0.901 & 0.654 & 0.796 & 0.493 & 0.472 & 0.592 &-0.037 &-0.007 & 0.398\\
  mFLAG-DR    & 0.833 & 0.534 & 0.907 & 0.928 & 0.805 & 0.906 & 0.672 & 0.439 & 0.563 &-0.203 &-0.119 & 0.482\\
  mFLAG-BT    & 0.520 & 0.578 & 0.790 & 0.827 & 0.424 & 0.627 & 0.668 & 0.431 & 0.556 &-0.166 &-0.100 & 0.494\\
  \midrule
  \multicolumn{13}{c}{Sarcasm$\rightarrow$Simile}\\
  \hline
  BART-Single & 0.585 & 0.163 & 0.897 & 0.906 & 0.666 & 0.793 & 0.276 & 0.460 & 0.581 &-0.091 &-0.056 & 0.241\\
  BART-Multi  & 0.588 & 0.003 & 0.911 & 0.932 & 0.725 & 0.857 & 0.006 & 0.471 & 0.594 &-0.050 &-0.013 & 0.005\\
  PT-to-FT    & 0.459 & 0.003 & 0.842 & 0.874 & 0.565 & 0.730 & 0.006 & 0.465 & 0.587 &-0.042 &-0.008 & 0.006\\
  mFLAG-DR    & 0.857 & 0.235 & 0.932 & 0.937 & 0.835 & 0.870 & 0.375 & 0.452 & 0.566 &-0.191 &-0.144 & 0.309\\
  mFLAG-BT    & 0.599 & 0.344 & 0.821 & 0.822 & 0.424 & 0.544 & 0.485 & 0.433 & 0.547 &-0.189 &-0.171 & 0.383\\
\bottomrule
\end{tabular}}
\caption{\label{tab:results-sarcasm}
 Results of sarcasm$\rightarrow$others generation. }
\end{table*}

\clearpage

\begin{table*}[!t]
\resizebox{\linewidth}{!}{%
\centering
\footnotesize
\begin{tabular}{l|cc|ccccc|ccccc}
\toprule
  & \multicolumn{2}{c|}{\textbf{Form Strength}} & \multicolumn{5}{c|}{\textbf{Source Text}} & \multicolumn{5}{c}{\textbf{Literal Text}} \\
  \cline{2-13}
  & \textbf{SRC} & \textbf{TGT} & \textbf{BLEU} & \textbf{BERT} & \textbf{BLEURT} & \textbf{COMET} & \textbf{HM} & \textbf{BLEU} & \textbf{BERT} & \textbf{BLEURT} & \textbf{COMET} & \textbf{HM}\\
  \midrule
  \multicolumn{13}{c}{Metaphor$\rightarrow$Hyperbole}\\
  \hline
  BART-Single & 0.173 & 0.480 & 0.617 & 0.786 & 0.446 & 0.582 & 0.540 & 0.588 & 0.779 & 0.399 & 0.511 & 0.529\\
  BART-Multi  & 0.260 & 0.427 & 0.643 & 0.826 & 0.562 & 0.722 & 0.513 & 0.635 & 0.825 & 0.561 & 0.700 & 0.511\\
  PT-to-FT    & 0.233 & 0.480 & 0.711 & 0.870 & 0.709 & 0.832 & 0.573 & 0.639 & 0.827 & 0.508 & 0.667 & 0.548\\
  mFLAG-DR    & 0.827 & 0.653 & 0.662 & 0.846 & 0.634 & 0.717 & 0.657 & 0.516 & 0.769 & 0.359 & 0.450 & 0.576\\
  mFLAG-BT    & 0.453 & 0.620 & 0.511 & 0.755 & 0.438 & 0.511 & 0.560 & 0.496 & 0.762 & 0.404 & 0.492 & 0.551\\
  \midrule
  \multicolumn{13}{c}{Metaphor$\rightarrow$Idiom}\\
  \hline
  BART-Single & 0.240 & 0.447 & 0.542 & 0.744 & 0.361 & 0.459 & 0.490 & 0.518 & 0.748 & 0.350 & 0.411 & 0.480\\
  BART-Multi  & 0.253 & 0.280 & 0.643 & 0.825 & 0.559 & 0.724 & 0.390 & 0.633 & 0.822 & 0.550 & 0.694 & 0.388\\
  PT-to-FT    & 0.113 & 0.260 & 0.646 & 0.819 & 0.573 & 0.748 & 0.371 & 0.657 & 0.834 & 0.554 & 0.683 & 0.373\\
  mFLAG-DR    & 0.887 & 0.547 & 0.640 & 0.829 & 0.582 & 0.708 & 0.590 & 0.542 & 0.787 & 0.444 & 0.561 & 0.544\\
  mFLAG-BT    & 0.653 & 0.547 & 0.557 & 0.771 & 0.453 & 0.586 & 0.552 & 0.524 & 0.774 & 0.416 & 0.541 & 0.536\\
  \midrule
  \multicolumn{13}{c}{Metaphor$\rightarrow$Sarcasm}\\
  \hline
  BART-Single & 0.133 & 0.240 & 0.623 & 0.788 & 0.424 & 0.604 & 0.347 & 0.597 & 0.782 & 0.391 & 0.532 & 0.347\\
  BART-Multi  & 0.233 & 0.280 & 0.654 & 0.820 & 0.527 & 0.712 & 0.392 & 0.621 & 0.807 & 0.510 & 0.652 & 0.386\\
  PT-to-FT    & 0.153 & 0.267 & 0.683 & 0.832 & 0.574 & 0.761 & 0.384 & 0.645 & 0.812 & 0.462 & 0.650 & 0.378\\
  mFLAG-DR    & 0.720 & 0.347 & 0.788 & 0.883 & 0.760 & 0.843 & 0.482 & 0.557 & 0.767 & 0.377 & 0.486 & 0.428\\
  mFLAG-BT    & 0.273 & 0.427 & 0.511 & 0.732 & 0.322 & 0.496 & 0.465 & 0.516 & 0.742 & 0.334 & 0.500 & 0.467\\
  \midrule
  \multicolumn{13}{c}{Metaphor$\rightarrow$Simile}\\
  \hline
  BART-Single & 0.107 & 0.087 & 0.631 & 0.785 & 0.418 & 0.574 & 0.153 & 0.598 & 0.775 & 0.384 & 0.489 & 0.152\\
  BART-Multi  & 0.273 & 0.007 & 0.647 & 0.828 & 0.569 & 0.733 & 0.014 & 0.637 & 0.826 & 0.579 & 0.710 & 0.014\\
  PT-to-FT    & 0.087 & 0.007 & 0.643 & 0.808 & 0.540 & 0.711 & 0.014 & 0.650 & 0.822 & 0.503 & 0.661 & 0.014\\
  mFLAG-DR    & 0.747 & 0.500 & 0.696 & 0.827 & 0.479 & 0.554 & 0.581 & 0.450 & 0.710 & 0.099 & 0.142 & 0.474\\
  mFLAG-BT    & 0.167 & 0.633 & 0.428 & 0.679 & 0.102 & 0.142 & 0.511 & 0.447 & 0.695 & 0.115 & 0.135 & 0.524\\
\bottomrule
\end{tabular}}
\caption{\label{tab:results-metaphor}
 Results of metaphor$\rightarrow$others generation. }
\end{table*}

\begin{table*}[!ht]
\resizebox{\linewidth}{!}{%
\centering
\footnotesize
\begin{tabular}{l|cc|ccccc|ccccc}
\toprule
  & \multicolumn{2}{c|}{\textbf{Form Strength}} & \multicolumn{5}{c|}{\textbf{Source Text}} & \multicolumn{5}{c}{\textbf{Literal Text}} \\
  \cline{2-13}
  & \textbf{SRC} & \textbf{TGT} & \textbf{BLEU} & \textbf{BERT} & \textbf{BLEURT} & \textbf{COMET} & \textbf{HM} & \textbf{BLEU} & \textbf{BERT} & \textbf{BLEURT} & \textbf{COMET} & \textbf{HM}\\
  \midrule
   \multicolumn{13}{c}{Simile$\rightarrow$Hyperbole}\\
  \hline
  BART-Single & 0.093 & 0.713 & 0.492 & 0.575 &-0.358 &-0.358 & 0.582 & 0.603 & 0.656 &-0.135 &-0.127 & 0.653\\
  BART-Multi  & 0.007 & 0.293 & 0.634 & 0.689 &-0.040 &-0.045 & 0.401 & 0.770 & 0.821 & 0.261 & 0.418 & 0.424\\
  PT-to-FT    & 0.000 & 0.327 & 0.649 & 0.692 & 0.003 &-0.012 & 0.435 & 0.777 & 0.818 & 0.261 & 0.417 & 0.460\\
  mFLAG-DR    & 0.527 & 0.893 & 0.895 & 0.918 & 0.772 & 0.811 & 0.894 & 0.583 & 0.685 &-0.041 &-0.090 & 0.705\\
  mFLAG-BT    & 0.240 & 0.820 & 0.640 & 0.687 &-0.035 &-0.022 & 0.719 & 0.657 & 0.756 & 0.162 & 0.171 & 0.730\\
  \midrule
  \multicolumn{13}{c}{Simile$\rightarrow$Idiom}\\
  \hline
  BART-Single & 0.127 & 0.627 & 0.488 & 0.554 &-0.367 &-0.440 & 0.549 & 0.589 & 0.646 &-0.169 &-0.204 & 0.607\\
  BART-Multi  & 0.007 & 0.207 & 0.634 & 0.689 &-0.040 &-0.045 & 0.273 & 0.770 & 0.821 & 0.261 & 0.418 & 0.326\\
  PT-to-FT    & 0.000 & 0.173 & 0.644 & 0.684 & 0.007 &-0.038 & 0.273 & 0.781 & 0.830 & 0.307 & 0.470 & 0.283\\
  mFLAG-DR    & 0.420 & 0.800 & 0.810 & 0.848 & 0.508 & 0.554 & 0.805 & 0.600 & 0.710 & 0.013 &-0.025 & 0.686\\
  mFLAG-BT    & 0.200 & 0.773 & 0.617 & 0.683 &-0.018 &-0.009 & 0.686 & 0.636 & 0.761 & 0.170 & 0.212 & 0.698\\
  \midrule
  \multicolumn{13}{c}{Simile$\rightarrow$Sarcasm} \\
  \hline
  BART-Single & 0.007 & 0.440 & 0.479 & 0.572 &-0.402 &-0.420 & 0.459 & 0.618 & 0.704 &-0.113 & 0.001 & 0.514\\
  BART-Multi  & 0.007 & 0.233 & 0.611 & 0.671 &-0.070 &-0.086 & 0.337 & 0.748 & 0.806 & 0.252 & 0.396 & 0.355\\
  PT-to-FT    & 0.000 & 0.387 & 0.551 & 0.623 &-0.128 &-0.178 & 0.455 & 0.677 & 0.743 & 0.117 & 0.242 & 0.492\\
  mFLAG-DR    & 0.373 & 0.367 & 0.877 & 0.892 & 0.671 & 0.692 & 0.517 & 0.619 & 0.714 & 0.001 &-0.014 & 0.598\\
  mFLAG-BT    & 0.073 & 0.380 & 0.618 & 0.672 &-0.057 &-0.057 & 0.471 & 0.726 & 0.792 & 0.241 & 0.362 & 0.499\\
  \midrule
  \multicolumn{13}{c}{Simile$\rightarrow$Metaphor}\\
  \hline
  BART-Single & 0.000 & 0.647 & 0.418 & 0.536 &-0.497 &-0.499 & 0.508 & 0.541 & 0.660 &-0.222 &-0.083 & 0.589\\
  BART-Multi  & 0.007 & 0.353 & 0.638 & 0.694 &-0.022 &-0.026 & 0.455 & 0.772 & 0.824 & 0.273 & 0.429 & 0.484\\
  PT-to-FT    & 0.000 & 0.367 & 0.643 & 0.685 &-0.007 &-0.041 & 0.467 & 0.782 & 0.825 & 0.289 & 0.445 & 0.500\\
  mFLAG-DR    & 0.440 & 0.680 & 0.815 & 0.878 & 0.595 & 0.702 & 0.741 & 0.552 & 0.681 &-0.036 &-0.099 & 0.609\\
  mFLAG-BT    & 0.013 & 0.773 & 0.550 & 0.638 &-0.167 &-0.166 & 0.643 & 0.668 & 0.757 & 0.079 & 0.256 & 0.717\\
\bottomrule
\end{tabular}}
\caption{\label{tab:results-simile}
 Results of simile$\rightarrow$others generation. }
\end{table*}

\clearpage
\subsection{Example Outputs of mFLAG}
\label{app:outputs}

\begin{table*}[!ht]
\footnotesize
\resizebox{\linewidth}{!}{%
\centering\begin{tabular}{l|p{10cm}|c}
\toprule
 \textbf{Forms} & \textbf{Sentences} & \textbf{Suc}.\\
 \midrule
 Literal [Input] & Old mr. smith has been teaching here for a very long time. & -\\ 
 Hyperbole & Old mr. smith has been teaching here since \textcolor{red}{the stone age}. & \textcolor{red}{\CheckmarkBold} \\
 \midrule
 Literal [Input] & I like your new haircut! & -\\
 Hyperbole & I love your new haircut! & \XSolidBrush \\
 \midrule
 Hyperbole [Input] & I am not happy that he urged me to finish all the hardest tasks in the world.& - \\
 Idiom & I am not happy that he urged me to \textcolor{red}{tick off} all the hardest tasks in the world. & \textcolor{red}{\CheckmarkBold} \\
 Sarcasm & \textcolor{red}{Glad} he urged me to finish all the hardest tasks in the world. & \textcolor{red}{\CheckmarkBold} \\
 Metaphor & I am not happy that he urged me to \textcolor{red}{bear} all the hardest tasks in the world. & \textcolor{red}{\CheckmarkBold} \\
 Simile & I am not happy that he urged me to finish all the \textcolor{red}{like a million things}. & \textcolor{red}{\CheckmarkBold} \\
 \midrule
 Idiom [Input] & Your exam results this term were not up to the mark.\\
 Hyperbole & Your exam results this term \textcolor{red}{were way over the top}. & \textcolor{red}{\CheckmarkBold}\\
 Sarcasm & Your exam results this term \textcolor{red}{were up to the mark}. & \textcolor{red}{\CheckmarkBold}\\
 Metaphor & Your exam results this term were not up to the mark. & \XSolidBrush\\
 Simile & Your exam results this term were not up to the mark. & \XSolidBrush\\
 \midrule
 Sarcasm [Input] & Being stuck in an airport is fun. & - \\
 Hyperbole & Being stuck in an airport is \textcolor{red}{the best}. & \textcolor{red}{\CheckmarkBold} \\
 Idiom & Being stuck in an airport is always fun. & \XSolidBrush\\
 Metaphor & Being stuck in an airport is no fun. & \XSolidBrush \\
 Simile & Being stuck in an airport is \textcolor{red}{like a rollercoaster}. & \textcolor{red}{\CheckmarkBold} \\
 \midrule
 Metaphor & The bad news rocked her hopes. & -\\
 Hyperbole & The bad news \textcolor{red}{crushed her hopes}. & \textcolor{red}{\CheckmarkBold}\\
 Idiom & The bad news crushed her hopes. & \XSolidBrush\\
 Sarcasm & The \textcolor{red}{good news} just keeps on \textcolor{red}{getting better}. & \textcolor{red}{\CheckmarkBold}\\
 Simile & The bad news rocked her \textcolor{red}{like a rock}. & \textcolor{red}{\CheckmarkBold}\\
  \midrule
 Simile [Input] & One day you decide to test it out and what you experience is like a magic trick. & - \\
 Hyperbole & One day you decide to test it out and  \textcolor{red}{what you experience is magic}. & \textcolor{red}{\CheckmarkBold}\\
 Sarcasm & One day you decide to test it out and  \textcolor{red}{what you experience is awesome}. & \textcolor{red}{\CheckmarkBold}\\
 Idiom & One day you decide to test it out and what you experience is dangerous. & \XSolidBrush \\
 Metaphor & One day you decide to test it out and what you experience is dangerous. & \XSolidBrush\\
\bottomrule
\end{tabular}}
\caption{\label{tab:example-outputs}
 Example outputs generated by mFLAG-DR, where \textcolor{red}{red} denotes appropriate words for desired forms.  Suc.==Successful.}
\end{table*}

\end{document}